\documentclass{bmvc2k}
\usepackage{graphicx}
\usepackage{comment}
\usepackage{amsmath,amssymb} 
\usepackage{color}
\usepackage{booktabs}
\usepackage{subfigure}
\usepackage{times}
\usepackage{epsfig}
\usepackage{mathrsfs}
\usepackage{url}

\title{ASAP-Net: Attention and Structure Aware Point Cloud Sequence Segmentation} 

\addauthor{Hanwen Cao *}{mbd_chw@sjtu.edu.cn}{1}
\addauthor{Yongyi Lu \dag}{yylu1989@gmail.com}{2}
\addauthor{Cewu Lu}{lucewu@sjtu.edu.cn}{1}
\addauthor{Bo Pang}{pangbo@sjtu.edu.cn}{1}
\addauthor{Gongshen Liu}{lgshen@sjtu.edu.cn}{1}
\addauthor{Alan Yuille}{alan.l.yuille@gmail.com}{2}

\addinstitution{
 Shanghai Jiaotong University\\
 Shanghai, China
}
\addinstitution{
 Johns Hopkins University\\
 Baltimore, MD, USA \\
 \quad \\
 * Work done at Johns Hopkins Univ. \\
 \dag corresponding author 
}

\runninghead{Cao Et al}{ASAP-Net for Point Cloud Sequence Segmentation}

\begin{document}

\maketitle

\begin{abstract}
Recent works of point clouds show that mulit-frame spatio-temporal modeling outperforms single-frame versions by utilizing cross-frame information. In this paper, we further improve spatio-temporal point cloud feature learning with a flexible module called ASAP considering both attention and structure information across frames, which we find as two important factors for successful segmentation in dynamic point clouds. Firstly, our ASAP module contains a novel attentive temporal embedding layer to fuse the relatively informative local features across frames in a recurrent fashion. Secondly, an efficient spatio-temporal correlation method is proposed to exploit more local structure for embedding, meanwhile enforcing temporal consistency and reducing computation complexity. Finally, we show the generalization ability of the proposed ASAP module with different backbone networks for point cloud sequence segmentation. Our ASAP-Net (backbone plus ASAP module) outperforms baselines and previous methods on both Synthia and SemanticKITTI datasets (+\textbf{3.4} to +\textbf{15.2} mIoU points with different backbones). 
Code is availabe at \url{https://github.com/intrepidChw/ASAP-Net}


\end{abstract}

\section{Introduction}
\label{introduction}
Dynamic point cloud sequences are readily-available input sources for many vision tasks. The ability to segment dynamic point clouds is a fundamental part of the perception system and will have a significant impact on applications such as autonomous driving \cite{azam2018object}, robot navigation \cite{salah2017summarizing} and augmented reality \cite{kung2016efficient}. 

\begin{figure}[h]
	\begin{center}
		\includegraphics[width=0.95\linewidth]{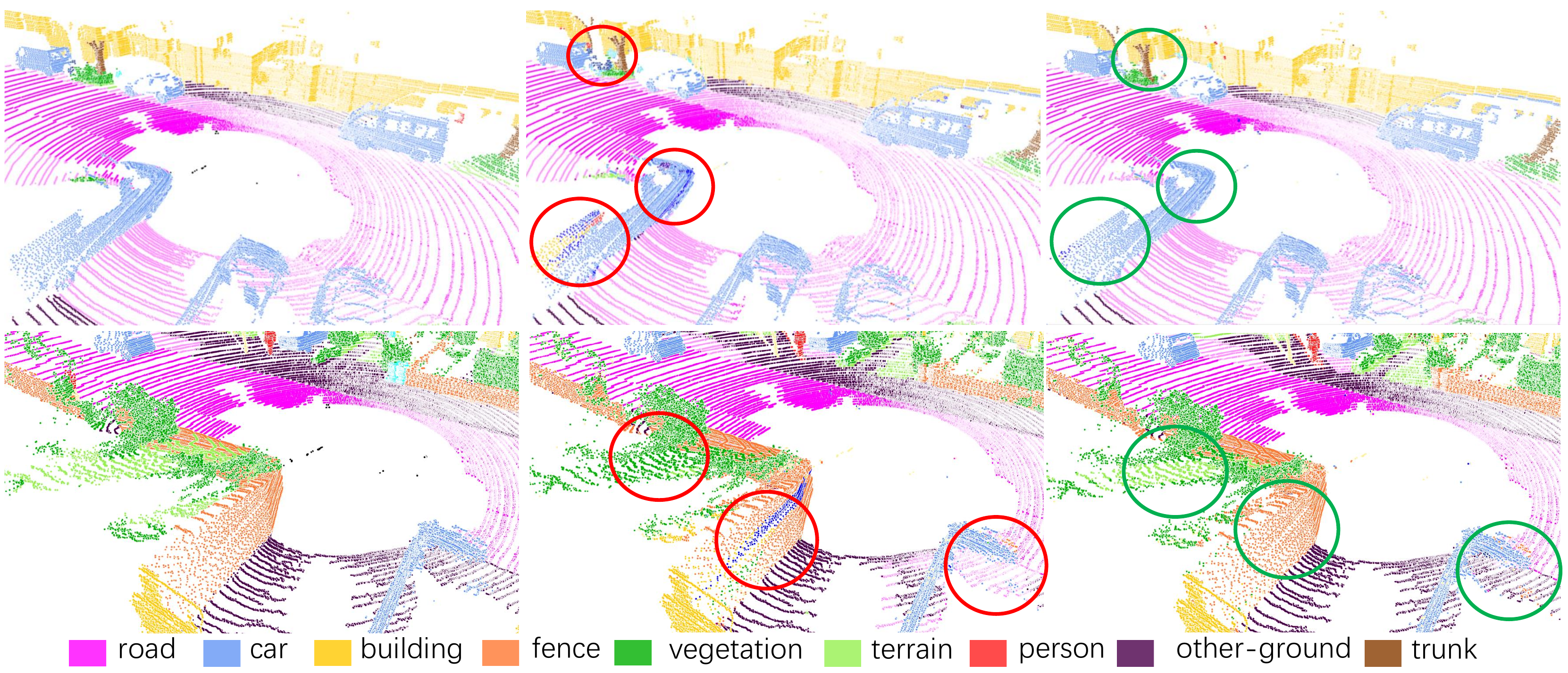} 
	\end{center}
	\vspace{-3mm}
	\caption{Segmentation results on SemanticKITTI~\cite{semanticKITTI}. From left to right: groundtruth, SqueezeSegV2~\cite{SqueezeSegV2} before and after incorporating our ASAP module. {\color{red}RED} circles highlight the failure cases by SqueezeSegV2 and {\color{green}GREEN} circles highlight the successful cases. Backbone's performance gets improved after incorporating our ASAP module.}
	\label{fig:teaser}
\end{figure}

While great success has been achieved in static point clouds~\cite{SqueezeSegV2,PointNet++,PointNet,SEGCloud,KnowNeighbors}, the literature on modeling point cloud sequence has not been fully-explored. For static point clouds, the pioneer work PointNet~\cite{PointNet} has shown its capability of directly learning point-wise features from the raw point clouds. However, a single point does not have semantic meaning until it is considered with its neighbouring points. Subsequent works on static point clouds~\cite{PointNet++,SEGCloud,KnowNeighbors} encode local structures from set of neighboring points and capture interactions among local structures. For modeling dynamic point cloud sequence, an intuitive and straightforward way is to encode the local structures both spatially and temporally, \emph{i.e.}, within a spatio-temporal neighborhood. Latest works~\cite{4DConv, meteornet} follow this direction. 
4D MinkNet~\cite{4DConv} first transforms the point cloud into regular voxels and applies a generalized sparse convolution along spatial and temporal dimensions. MeteorNet~\cite{meteornet} directly stacks multi-frame point clouds and computes local features by grouping spatio-temporal neighboring points. Although both methods are capable of capturing spatio-temporal features, 
it is hard for them to decouple spatial and temporal information in sequences~\cite{deepRNN,nature}. There is evidence that in real world, the human visual system largely relies on temporal information~\cite{lowvision} and treats it as a primary source of information. Without decoupling spatial and temporal structures, it is not well understood to which degree temporal information can contribute to dynamic scene understanding~\cite{ValueOfTem}. Therefore, it will be beneficial to related research if we can decouple the two aspects of information.
Also, decoupling spatial and temporal structures can make the model more extensible and flexible, \emph{i.e.}, we can easily apply the static point cloud methods as backbone networks, which can be seen in Section~\ref{ASAP module} and Section~\ref{implementation}. 
%


To effectively encode temporal structure, we need to tackle the following challenges:

(a) \textbf{Feature fusion with different frames}: The features from different frames may contribute differently to the results and they are all likely to contain undesired noise or mistakes. Ideally, the network should automatically identify the importance or confidence degree of different frames to achieve better fusing results. 

(b) \textbf{Point correlation across frames}: To fuse features from different frames, we need to correlate points across frames. However, the distribution of dynamic points varies from time to time and they are unordered, making it challenging to correlate.

In our ASAP module, we propose a novel attentive temporal embedding layer and spatio-temporal correlation strategy to tackle the above two challenges respectively. We also conduct thorough experiments to show our ASAP module's advantage over state-of-the-art methods~\cite{4DConv, meteornet} and its generalization ability of improving the performance of different backbones. Figure~\ref{fig:teaser} provides qualitative results of our approach, showing the effectiveness of our ASAP module. Our key contributions are:

\begin{itemize}
    \vspace{-3mm}
	\item We run a pilot study towards segmentation in point cloud sequences, by proposing a flexible architecture called ASAP module which can be easily plugged into previous static point cloud pipeline and achieve improvement by a large margin. 
	\vspace{-2mm}
	\item We introduce a novel attentive temporal embedding layer to fuse the spatial local features across frames effectively and robustly by automatically calculating attentions. 
	\vspace{-6mm}
	\item We present a spatio-temporal correlation strategy to exploit structural information, enforce temporal consistency and reduce computation.
\end{itemize}

\section{Related Work}

\subsection{Semantic Segmentation in Static Point Clouds}
\label{RelatedWokrk1}

Semantic segmentation in point clouds is a long-standing topic. 
Deep neural networks designed for the task can be divided into four categories:
a) \textbf{Voxelization based methods}. 
These methods transform the point cloud into regular voxel grids and apply 3D convolution network to do semantic segmentation \cite{PointLabeling}. However, the sparsity of point clouds can cause exploding memory consumption problem. More advanced methods~\cite{SEGCloud, OctNet, VoxSegNet, SSCNN} are proposed to tackle the problem.
b) \textbf{Range image based methods}.
These methods \cite{SqueezeSeg,SqueezeSegV2,RIUNet} first project the point cloud onto sphere according to the azimuth and zenith angles and then apply traditional convolutional networks like SqueezeNet \cite{SqueezeNet} and U-Net \cite{UNet}.
c) \textbf{Direct point cloud methods.}
These methods directly take point clouds as input. \cite{PointNet,PointNet++} extract point-wise features using multi-layer perceptron and symmetric function (max pooling) to aggregate features. \cite{PointSIFT} proposes an orientation-encoding unit to improve PointNet-based architectures' representation ability. \cite{KnowNeighbors} proposes new grouping techniques to better define neighborhoods for PointNet-based feature network.
\cite{PointwiseConv,FlexConvolution,TangentConv} propose new convolution operators to directly deal with point clouds.
d) \textbf{Other representation based methods}.
\cite{EscapeFromCells,KdTreeGuided} use kd-trees to represent point clouds.
\cite{RGCNN,DynamicGraph,SPGraph} are graph-based.
\cite{SPLATNet} interpolates input features onto a permutohedral lattice then does convolution.

\subsection{Temporal Information in Point Cloud Sequences}
To deal with 3D-videos (a continuous sequence of depth images, or LiDAR scans), \cite{4DConv} adopts sparse tensors and proposes the generalized sparse convolution. 
\cite{meteornet} is based on PointNet++~\cite{PointNet++}. It first stacks different frames of point clouds and get neighbor points in different frames by either searching with a time-dependent radius or tracking the points.

Other methods include scene flow and motion estimation. \cite{RigidSceneFlow} formulates the problem as an energy minimization problem of a factor graph, with hand-crafted SHOT \cite{SHOT} descriptors for correspondence search. \cite{LearnigApproachSceneFlow} uses EM algorithm to estimate a locally rigid and non-deforming flow. \cite{FlowNet3D} is based on PointNet \cite{PointNet} and PointNet++ \cite{PointNet++} and proposes a new flow embedding layer and set upconv layer to estimate scene flow in point clouds in an end-to-end way. \cite{PointFlowNet} estimates scene flow as rigid motions of individual objects or background with network that jointly learns to regress ego-motion and detect 3D objects. All these methods focus on different tasks and can not be directly applied to the task of semantic segmentaion.

\section{ASAP-Net}

In this section, we will present our ASAP module and the overall architecture.

\subsection{ASAP module}
\label{ASAP module}

\begin{figure}[t]
	\begin{center}
		\includegraphics[width=1.0\linewidth]{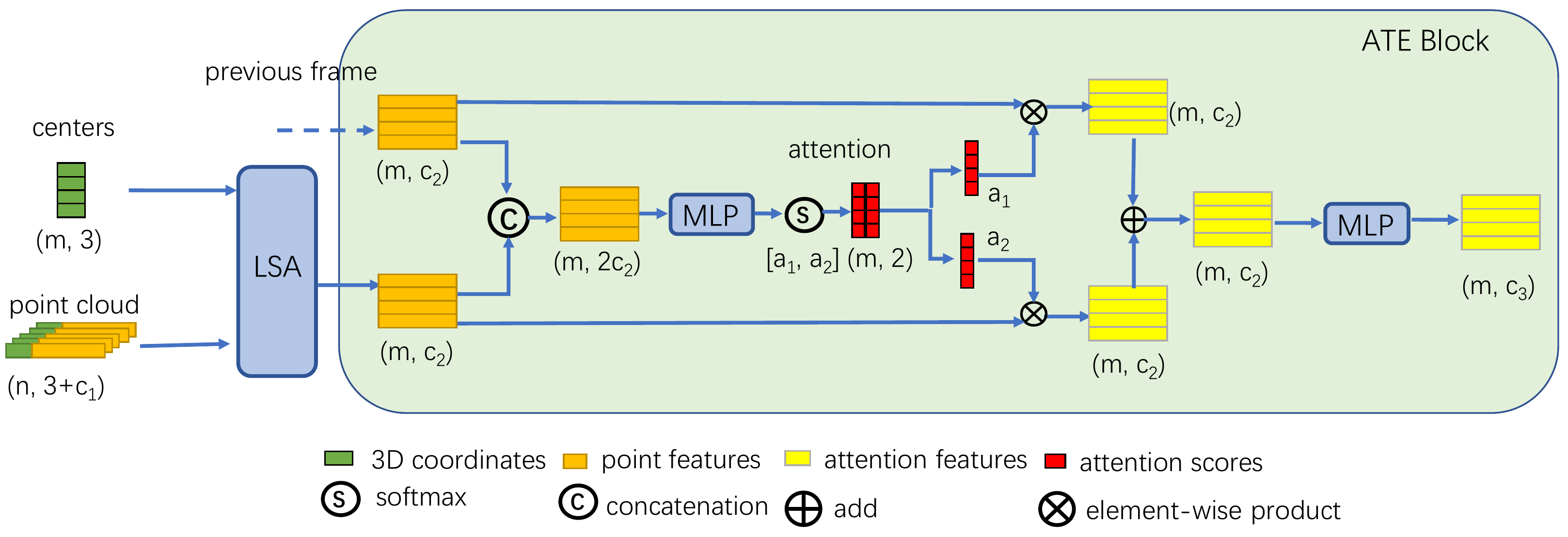} 
	\end{center}
	\vspace{-5mm}
	\caption{The proposed ASAP module. Given the current point cloud and corresponding center points, we first use the LSA block to compute the center features. We then fuse the center features from consecutive frames with the ATE block. The correspondence between the two sets of center features are guaranteed by our spatio-temporal correlation strategy.}
	\label{fig:ASAP}
\end{figure}

The architecture of our ASAP module is illustrated in Figure~\ref{fig:ASAP}. To illustrate our module, we denote a point cloud sequence of length $T$ as $\{S_1, S_2, ..., S_T\}$. Each frame of point cloud is represented as $S_t = \{p_i^{(t)}\}^{n}_{i=1}$, where $n$ is the number of points and each point $p_i^{(t)}$ consists of its Euclidean coordinate $x_i^{(t)} \in \mathbb{R}^3 $ and feature $f_i^{(t)} \in \mathbb{R}^c$. We also denote a set of center points in $S_t$ as $\{c_j^{(t)}\}^{m}_{j=1}$, where $m$ is the number of center points and $m<n$. 

\subsubsection{Local Structure Aggregation (LSA)}

Given a single frame of point cloud represented as $\{p_i^{(t)}\}^{n}_{i=1}$ and corresponding center points $\{c_j^{(t)}\}^{m}_{j=1}$, we follow PointNet++ \cite{PointNet++} to extract local features. For each center point, we gather its neighbor points within a radius, denoted as $\mathcal{N}(c_j^{(t)})$, and compute its feature with the following symmetric function:

\begin{equation}
\label{pointnet_frame1}
f_j^{(t)} = \mathop{MAX} \limits_{p_i^{(t)} \in \mathcal{N}(c_j^{(t)})} \{ \eta (f_i^{(t)}, x_i^{(t)}-x_j^{(t)})\}.
\end{equation}
where $\eta$ is an MLP (multi-layer perceptron) with concatenated feature vector $f_i^{(t)}$ and difference of spatial positions $x_i^{(t)}-x_j^{(t)}$ as inputs. $MAX$ is element-wise max pooling. 


\subsubsection{Spatio-temporal Correlation (STC)}
\label{SC}
Recall that we need to tackle two main challenges in dynamic point clouds: (a) Feature fusion with different frames and (b) point correlation across frames. Here we first introduce (b) how to correlate points between frames and leave (a) in the next section. Specifically, given two consecutive point clouds, we need to obtain pairwise correspondence between them to leverage temporal information. Latest methods like \cite{meteornet} use either direct grouping of neighboring points or flow-based grouping. The former is incapable of knowing exactly the correspondences across frames,
while the latter are computationally expensive. In the followings we present two simple yet efficient correlation approaches:

(i) \textbf{Per-frame Nearest Center Search}. Use farthest point sampling to generate center points each frame and compute local features separately. For each center in the current frame, we correlate it with the closest center in the previous frame based on Euclidean distance.   

(ii) \textbf{Inter-frame Constant Center Iteration}. Only generate the center coordinates in the first frame using farthest point sampling and use the same center coordinates in subsequent frames so the center features are naturally correlated.

\begin{figure}[t]
	\begin{center}
		\includegraphics[width=0.95\linewidth]{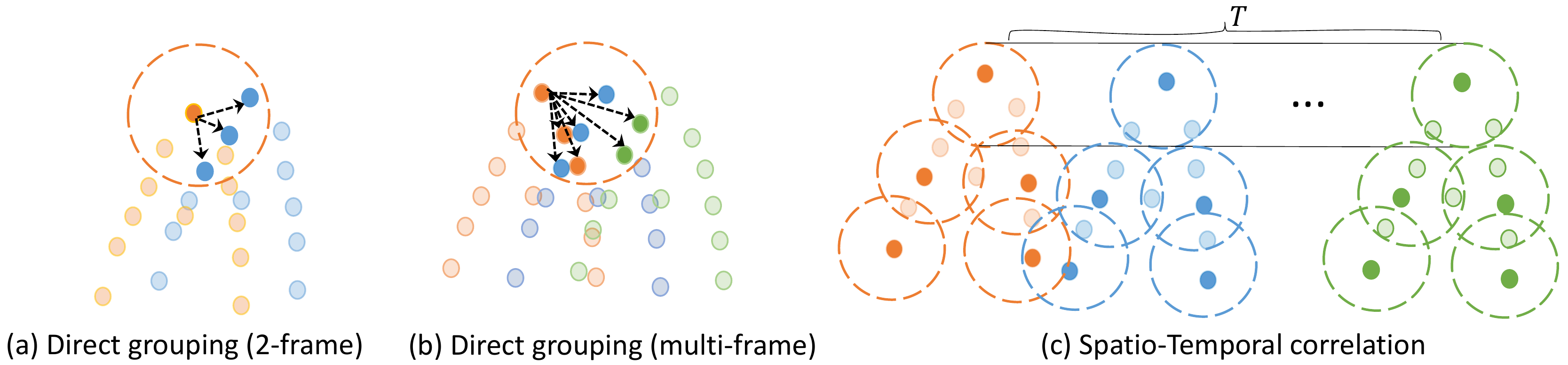} 
	\end{center}
	\vspace{-5mm}
	\caption{Illustration of the proposed spatio-temporal correlation. Different from (a) two-frame direct grouping \cite{FlowNet3D} and (b) multi-frame direct grouping \cite{meteornet}, each center (the solid points inside the cycles in (c)) in our spatio-temporal correlation are correlated across $T$ frames either by (i) per-frame nearest center search or (ii) inter-frame constant center iterate, followed by the temporal embedding layer. Compared to (a) and (b), our tubular-alike correlation is able to decouple spatial and temporal structures.}
	\label{fig:p2p}
\end{figure}

Our insight to design (ii) is that, to capture temporal information effectively, the network should have stable focus regions to achieve temporal consistency. It may seem uncommon to use the same center coordinates since they don't exist in frames except the first one. However, their neighbor regions do exist. Although the points are dynamic, their boundary is actually stable (defined by the intrinsic parameters of LiDAR scanner or depth camera). Therefore, there's actually no worry that some center points get too far away and no longer have neighbor points due to the dynamics the point cloud. Each center point forms a spatio-temporal tube in time (see Figure \ref{fig:p2p}) and by computing the center features in a recurrent way, we can capture temporal information and adapt to the neighborhood changes.

We adopt (ii) as our final strategy. It can be seen that (i) needs to do sampling per frame and compute the distances while (ii) is more computationally efficient. We will show the advantages of our spatio-temporal correlation strategy with experiments in Section~\ref{updatingExperiment}.


\subsubsection{Temporal Embedding (TE)}
\label{TE}
The temporal embedding layer is used to fuse local structure features across frames. The layer will take in the current center points $\{c_j^{(t)}\}^{m}_{j=1}$ and the matched previous ones $\{c_j^{(t-1)}\}^{m}_{j=1}$ after spatio-temporal correlation and update the feature of each current center point which we denote as $h(c_j^{(t)})$. We also propose two temporal embedding methods:

(i) \textbf{Direct Temporal Embedding (DTE)}: We directly concatenate the features of the matched center points and use a shared MLP $\zeta$ to update the feature of current center point: $h(c_j^{(t)}) = \zeta (f_j^{(t-1)}, f_j^{(t)})$.

(ii) \textbf{Attentive Temporal Embedding (ATE)}: We compute two scalar attentions by feeding the concatenation of two correlated features into a shared MLP $\gamma$ followed by a $Softmax$ function: $[a_1, a_2] = Softmax(\gamma (f_j^{(t-1)}, f_j^{(t)}))$.
Then we calculate an weighted-sum feature by multiply the two attentions with corresponding features: $f_j^{'(t)} = a_1f_j^{(t-1)}+ a_2f_j^{(t)}$.
Finally we get the updated feature of current center point with another MLP $\zeta$: $h(c_j^{(t)}) = \zeta (f_j^{'(t)})$.

Our insight to design the ATE block is that different frames may contribute differently to the results. For slowly-moving objects, features from previous frame can be more reliable since they contain multi-frame information due to the recurrent computation while for fast-moving objects, it's better to rely more on current frame. Also, there can be undesired mistakes and noise in each frame. Therefore, the network can make better and more robust predictions by estimating the importance and reliability of features from different frames. ATE can achieve better results with even less parameters, which will be shown in Section~\ref{experiments}.

The LSA and TE can be stacked hierarchically. Specifically, we call the module with DTE as SAP-x (no attention) and the module with ATE as ASAP-x (with attention), where x is the number of LSA and TE.

\begin{figure}[t]
	\begin{center}
		\includegraphics[width=0.95\linewidth]{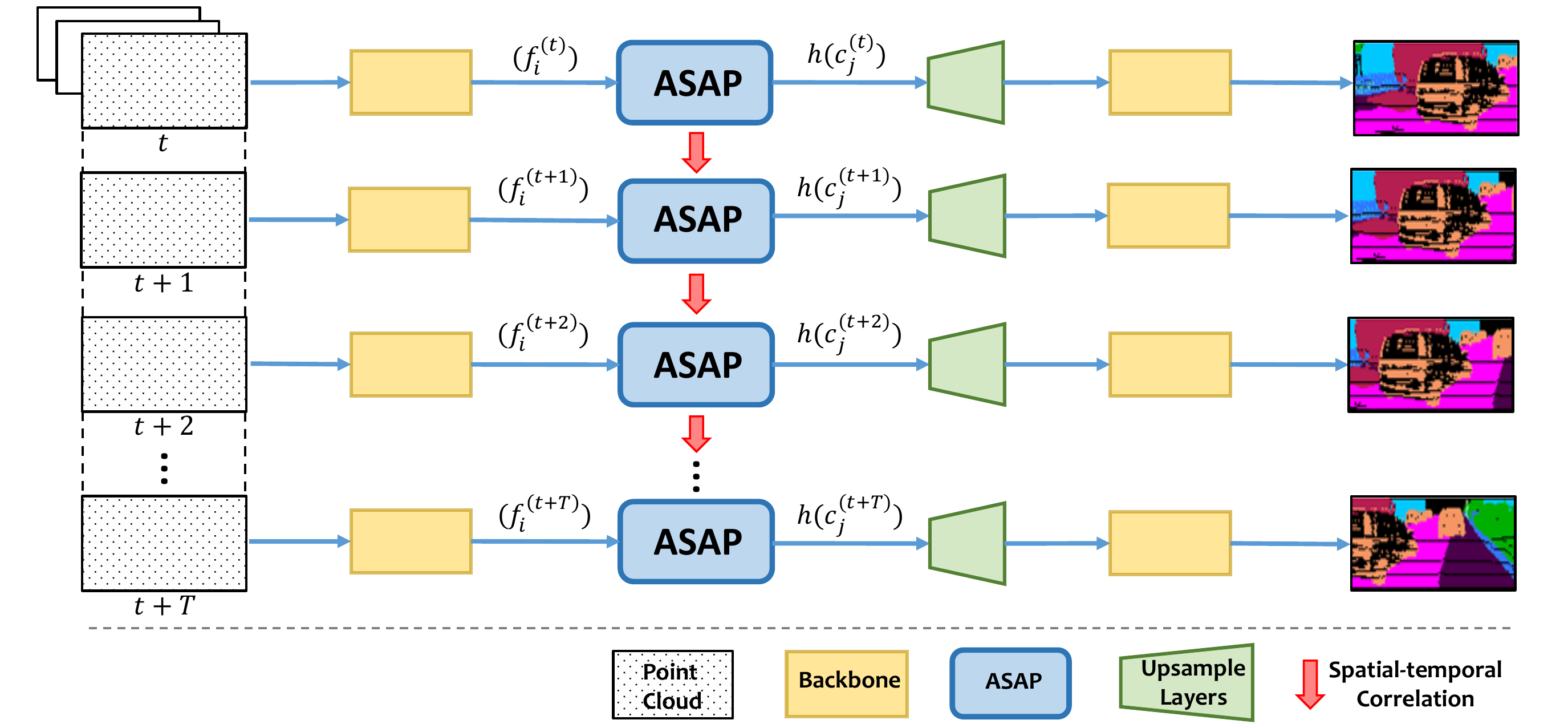} 
	\end{center}
	\vspace{-5mm}
	\caption{Overall network structure. Our framework consists of the backbone network and the proposed ASAP module and upsampling layers. We also use our spatio-temporal correlation strategy to guarantee the correlation between consecutive frames.} 
	\label{fig:network}
\end{figure}

\subsection{Overall Structure}

The overall structure is shown in Figure~\ref{fig:network}. We first use a backbone network like~\cite{PointNet++, SqueezeSegV2} to extract point features in each frame. Then our ASAP module will compute the center features and fuse across frames in a recurrent way. Next, we use the \textit{feature propagation} layer proposed by \cite{PointNet++} to upsample the point cloud to original size and return it to the backbone network. Finally, the backbone network will continue to run and return the predictions. 

According to our spatio-temporal correlation strategy, for each sequence, we sample center points with farthest point sampling in the first frame and use the same coordinates in subsequent frames to achieve temporal consistency and reduce computation.

\section{Experiments}
\label{experiments}

\subsection{Datasets and Settings}

We conduct semantic segmentation experiments on two datasets. We first test our method on the large-scale synthetic dataset Synthia \cite{synthia}, compare with meteornet \cite{meteornet} and a sparse 4D CNN \cite{4DConv} baselines and do ablation study. We then conduct experiments on the largest publicly available real LiDAR dataset semanticKITTI \cite{semanticKITTI} to show our module's effectiveness and generalization ability. 

\textbf{Synthia} It consists of six sequences of driving scenarios in nine different weather conditions. Each sequence consists of four stereo RGBD images from four viewpoints captured from the top of a moving car. We reconstruct 3D point clouds from RGBD images and create point cloud sequences the same way with \cite{meteornet}. We used the same train/validation/test split as \cite{meteornet,4DConv}. 
The train, validation and test set contain 19,888, 815 and 1,886 frames respectively.

\textbf{SemanticKITTI} 
The dataset provides 23,201 full 3D scans for training and 20,351 for testing, which makes it by a wide margin the largest dataset publicly available. We follow the split in \cite{semanticKITTI} for training, validation and testing. To compare with the backbone network, we follow the single scan experiments proposed in \cite{semanticKITTI} and make predictions for the 19 classes. Since our model take a sequence as input, we split the testing set into sequences with a fixed length and save the result of each scan.

\subsection{Choice of Backbone Networks}
\label{implementation}
Our proposed ASAP module is a flexible architecture and can be easily incorporated into different backbones. We call the overall network ASAP-Net (Backbone + ASAP module). 

To show the flexibility of our ASAP module, the backbones should be different in nature. Therefore, PointNet++\cite{PointNet++} and SqueezeSegV2~\cite{SqueezeSegV2} are chosen for the simplicity of architecture and representativeness in two different categories: 1) directly processing and 2) range image based, as per-frame point cloud methods mainly fall into these two groups. For PointNet++~\cite{PointNet++},  our ASAP module processes the point cloud in the same way so it can be incorporated directly; For SqueezeSegV2~\cite{SqueezeSegV2}, the network first projects the LiDAR point cloud onto a sphere for a dense, grid-based representation as $\theta = \arcsin \frac{z}{\sqrt{x^2 + y^2 + z^2}}, \tilde{\theta} = \lfloor \theta / \Delta \theta \rfloor$ and $\phi = \arcsin \frac{y}{\sqrt{x^2 + y^2}}, \tilde{\phi} = \lfloor \phi / \Delta \theta \rfloor$.
Here $\Delta \theta$ and $\Delta \phi$ are resolutions for discretization and $(\tilde{\theta},\tilde{\phi})$ denotes the position of a point on a 2D spherical grid. After applying these equations on each point cloud, we can obtain a 3D tensor of size $H \times W \times C$. Then they adopt an encoder-decoder structure using CNN module called \textit{fireModule} and \textit{fireDeconvs}, which are introduced in SqueezeNet \cite{SqueezeNet}. 

To reconstruct the point cloud from 2D spherical grid, we simply reshape the feature map of the middle layer from $H \times W \times C$ to $N \times C$, where $N = H \times W$ is the number of points and $C$ is the feature length. We use max pooling layer on the original projected spherical grid (the channel contains coordinates) to get the corresponding 3D coordinates approximately.

Note that, Although as claimed in \cite{semanticKITTI}, \textit{DarkNet53Seg} achieves the best mIoU result on SemanticKITTI \cite{semanticKITTI}, it's not a published work. According to the author, it's the same type with SqueezeSegV2~\cite{SqueezeSegV2} and it performs better just because it has more layers. Therefore, improvement on SqueezeSegV2~\cite{SqueezeSegV2} should guarantee the improvement on DarkNet53Seg.

\begin{table*}[t] 
	\scriptsize\centering
	\setlength{\tabcolsep}{0.45mm}{
		\begin{tabular}{@{}ccccccccccccccccc@{}}
			\toprule
			 & \multicolumn{1}{l|}{ param} & \multicolumn{1}{c|}{ frame} & & \multicolumn{1}{l|}{} & \multicolumn{12}{c}{IoU} \\
			Approach & \multicolumn{1}{c|}{ (M)} & \multicolumn{1}{c|}{num} & mIoU & \multicolumn{1}{l|}{mAcc} & 
			Bldg & Road & Sdwlk & Fence & Vegitn & Pole & Car & T.Sign & Pdstr & Bicyc & Lane & T.light \\ \midrule
			3D MinkNet \cite{4DConv} & \multicolumn{1}{c|}{19.31} & \multicolumn{1}{c|}{ 1} & 76.24 & \multicolumn{1}{l|}{\textbf{89.31}} & 89.39 & 97.68 & 69.43 & 86.52 & 98.11 & 97.26 & 93.50 & 79.45 & 92.27 & 0.00 & 44.61 & 66.69 \\
			4D MinkNet \cite{4DConv} & \multicolumn{1}{c|}{23.72} & \multicolumn{1}{c|}{ 3} & 77.46 & \multicolumn{1}{l|}{88.01} & 90.13 & \textbf{98.26} & 73.47 & 87.19 & \textbf{99.10} & 97.50 & 94.01 & 79.04 & \textbf{92.62} & 0.00 & 50.01 & 68.14  \\
			\midrule
			PointNet++ \cite{PointNet++} & \multicolumn{1}{c|}{0.88} & \multicolumn{1}{c|}{ 1 } & 79.35 & \multicolumn{1}{l|}{85.43} & 96.88 & 97.72 & 86.20 & 92.75 & 97.12 & 97.09 & 90.85 & 66.87 & 78.64 & 0.00 & 72.93 & 75.17  \\
			\midrule
			MeteorNet \cite{meteornet} & \multicolumn{1}{c|}{1.78} & \multicolumn{1}{c|}{ 3 } & 81.80 & \multicolumn{1}{l|}{86.78} & \textbf{98.10} & 97.72 & 88.65 & 94.00 & 97.98 & \textbf{97.65} & 93.83 & \textbf{84.07} & 80.90 & 0.00 & 71.14 & 77.60  \\
			PNv2 with SAP-1 & \multicolumn{1}{c|}{1.97} & \multicolumn{1}{c|}{ 3} & 82.31 & \multicolumn{1}{l|}{86.72} & 97.68 & 97.99 & \textbf{90.16} & 94.84 & 97.25 & 97.34 & 94.77 & 80.35 & 83.54 & 0.00 & \textbf{75.13} & \textbf{78.62}  \\
			PNv2 with ASAP-1 & \multicolumn{1}{c|}{1.84} & \multicolumn{1}{c|}{ 3} & \textbf{82.73} & \multicolumn{1}{l|}{87.02} & 97.67 & 98.15 & 89.85 & \textbf{95.50} & 97.12 & 97.59 & \textbf{94.90} & 80.97 & 86.08 & 0.00 & 74.66 & 77.51  \\
			\bottomrule
		\end{tabular}
	}
	\caption{Segmentation results on the Synthia dataset \cite{synthia}. The evaluation metrics are mean IoU and mean accuracy ($\%$). PNv2 represents PointNet++~\cite{PointNet++}. SAP-x and ASAP-x are defined in Section~\ref{TE}.}
	\label{tab:result_synthia}
\end{table*}

\subsection{Evaluation on Synthia}\label{results_synthia}

\begin{figure}[t]
	\begin{center}
		\includegraphics[width=0.95\linewidth]{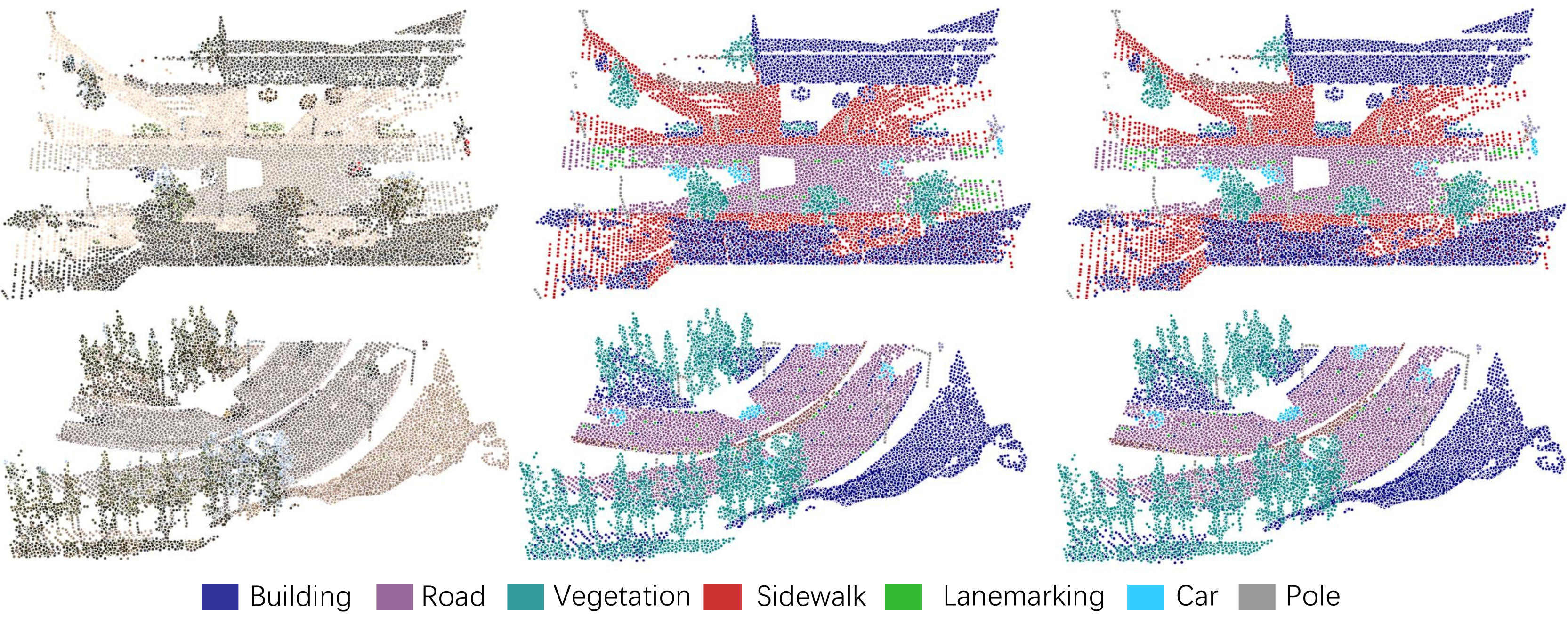} 
	\end{center}
	\vspace{-5mm}
	\caption{Visualization of two example results from the Synthia dataset. From left to right: RGB input, groundtruth, predictions.}
	\label{fig:synthia_results}
\end{figure}

The results are listed in Table~\ref{tab:result_synthia}, our ASAP-Net consistently surpasses both the sparse 4D CNN \cite{4DConv} and MeteorNet \cite{meteornet} and establishes a new state-of-art result in term of mIoU. Figure~\ref{fig:synthia_results} visualizes two samples. Our model can accurately segment most objects.

We can see that our ASAP module can achieve better results with even less parameters than SAP module. The reduction of parameters has two aspects of reasons. Firstly, the extra number of parameters $\gamma$ to compute attention is really small since the output attention length is just two. Secondly, the features from two consecutive frames are fused by addition instead of concatenation so the input feature length to $\zeta$ is half of that in SAP.

\subsection{Evaluation on SemanticKITTI}\label{results}

The IoU results of the backbone networks as well as the results with our ASAP-Net are shown in Table~\ref{tab:results}. The performances of both backbone networks get improved when combined with our ASAP-module. Some quantitative visualizations can be found in Figure~\ref{fig:teaser}. 

\begin{table*}[t] 
	\scriptsize\centering
	\setlength{\tabcolsep}{0.45mm}{
		\begin{tabular}{@{}lllllllllllllllllllll@{}}
			\toprule
			Approach & \multicolumn{1}{l|}{ \rotatebox{90}{mIoU}} & \rotatebox{90}{car}  & \rotatebox{90}{bicycle} & \rotatebox{90}{motorcycle} & \rotatebox{90}{truck} & \rotatebox{90}{other-vehicle} & \rotatebox{90}{person} & \rotatebox{90}{bicyclist} & \rotatebox{90}{motorcyclist} & \rotatebox{90}{road} & \rotatebox{90}{parking} & \rotatebox{90}{sidewalk} & \rotatebox{90}{other-ground} & \rotatebox{90}{building} & \rotatebox{90}{fence} & \rotatebox{90}{vegetation} & \rotatebox{90}{trunk} & \rotatebox{90}{terrain} & \rotatebox{90}{pole} & \rotatebox{90}{traffic-sign} \\ \midrule
			
			PointNet~\cite{PointNet} & \multicolumn{1}{l|}{14.6} & 46.3 & 1.3 & 0.3 & 0.1 & 0.8 & 0.2 & 0.2 & 0.0 & 61.6 & 15.8 & 35.7 & 1.4 & 41.4 & 12.9 & 31.0 & 4.6 & 17.6 & 2.4 & 3.7  \\

			PointNet++~\cite{PointNet++}& \multicolumn{1}{l|}{20.1} & 53.7 & 1.9 & 0.2 & 0.9 & 0.2 & 0.9 & 1.0 & 0.0 & 72.0 & 18.7 & 41.8 & 5.6 & 62.3 & 16.9 & 46.5 & 13.8 & 20.0 & 6.0 & 8.9  \\
			
			PNv2 with SAP-1 & \multicolumn{1}{l|}{31.4} & 79.7 & 9.5 & 5.5 & 0.1 & 8.1 & 7.9 & 22.3 & 1.1 & 81.2 & \textbf{34.2} & 56.3 & 7.9 & 75.7 & 38.5 & 58.8 & 26.8 & 50.1 & 16.1 & 16.9  \\
			
			PNv2 with ASAP-1 & \multicolumn{1}{l|}{33.3} & 84.1 & \textbf{11.6} & \textbf{7.5} & 3.2 & 11.4 & 7.8 & 18.5 & 3.0 & \textbf{81.8} & 28.1 & 53.1 & 7.8 & 74.9 & 37.6 & 64.4 & 27.2 & 51.7 & 22.8 & \textbf{30.8}  \\
			
			
			PNv2 with ASAP-2 & \multicolumn{1}{l|}{\textbf{35.3}} & \textbf{86.4} & 9.3 & 6.4 & \textbf{8.1} & \textbf{13.0} & \textbf{12.8} & \textbf{25.2} & \textbf{3.8} & 80.3 & 29.7 & \textbf{57.6} & \textbf{13.2} & \textbf{77.7} & \textbf{40.1} & \textbf{66.7} & \textbf{31.9} & \textbf{52.9} & \textbf{26.7} & 28.5  \\
			
			\midrule
			
			SqueezeSeg~\cite{SqueezeSeg} & \multicolumn{1}{l|}{29.5} &	68.8 &	16.0 &	4.1 &	3.3 &	3.6 &	12.9 &	13.1 &	0.9 &	85.4 &	26.9 &	54.3 &	4.5 &	57.4 &	29.0 &	60.0 &	24.3 &	53.7 &	17.5 &	24.5 \\
			
			SqueezeSegV2~\cite{SqueezeSegV2} & \multicolumn{1}{l|}{39.7} & 81.8 & 18.5    & 17.9       & 13.4  & 14.0          & 20.1   & 25.1      & 3.9          & 88.6 & 45.8    & 67.6     & 17.7         & 73.7     & 41.1  & 71.8       & 35.8  & 60.2    & 20.2 & 36.3         \\
			
			SSv2 with SAP-1 & \multicolumn{1}{l|}{40.7} & 83.0 & 21.1    & \textbf{22.4}       & 4.1   & 15.5          & 25.7   & \textbf{33.9}      & 1.3          & 89.5 & 52.9    & 68.9     & 19.2         & 73.7     & 40.1  & 69.2       & 36.1  & 61.4    & 14.0 & \textbf{40.9}         \\
			
			SSv2 with ASAP-1 & \multicolumn{1}{l|}{41.3} & 83.3 & 19.8    & 21.1       & 9.2   & 15.5          & 24.4   & 31.3      & 2.5          & 89.7 & 53.3    & 69.0     & 18.1         & 79.3     & 45.0  & 71.7       & \textbf{38.3}  & 61.6    & 14.0 & 37.1         \\
			
			SSv2 ASAP-1-msg & \multicolumn{1}{l|}{42.5} & \textbf{84.0} & 19.8    & 22.3       & 13.1   & 17.2          & 24.0   & 31.9      & \textbf{5.5}          & 90.4 & \textbf{55.5}    & \textbf{70.9}     & \textbf{22.2}         & 80.5     & 46.2  & 71.5       & 37.0  & 63.1    & 15.3 & 36.5         \\
			
			SSv2 with ASAP-2   & \multicolumn{1}{l|}{\textbf{43.1}} & 83.1 & \textbf{22.5}    & 20.8       & \textbf{16.0}  & \textbf{18.1}          & \textbf{26.6}   & 32.8      & 4.6          & \textbf{90.5} & 55.2    & 69.8     & 20.4     & \textbf{81.7}       & \textbf{49.0}  & \textbf{72.0}       & \textbf{38.3}  & \textbf{63.2}    & \textbf{16.4} & 38.4         \\
			
			\midrule
			
			SPGraph~\cite{SPGraph} & \multicolumn{1}{l|}{17.4} & 49.3 & 0.2 & 0.2 & 0.1 & 0.8 &	0.3 & 2.7 &	0.1 & 45.0 & 0.6 & 39.1 &	0.6 & 64.3 & 20.8 & 48.9 &	27.2 &	24.6 &	15.9 &	0.8 \\
			
			SPLATNet~\cite{SPLATNet} & \multicolumn{1}{l|}{18.4} &	58.2 &	0.0 &	0.0 &	0.0 &	0.0 &	0.0 &	0.0 &	0.0 &	64.6 &	0.4 &	39.1 &	0.0 &	58.3 &	23.1 &	71.1 &	9.9 &	19.3 &	5.6	& 0.0 \\
			
			TangentConv~\cite{TangentConv} & \multicolumn{1}{l|}{40.9} &	\textbf{90.8} &	2.7 &	16.5 &	15.2 &	12.1 &	23.0 &	28.4 &	\textbf{8.1} &	83.9 &	33.4 &	63.9 &	15.4 &	83.4 &	49.0 &	\textbf{79.5} &	49.3 &	58.1 &	35.8 &	28.5 \\
			
			DarkNet21Seg~\cite{semanticKITTI} & \multicolumn{1}{l|}{47.4} & 85.4 & \textbf{26.2} & 26.5 & 18.6 & 15.6 &	31.8 & \textbf{33.6} &	4.0 & 91.4 & 57.0 & 74.0 &	26.4 & 81.9 & 52.3 & 77.6 & 48.4 &	63.6 &	24.6 &	50.0 \\
			
			DarkNet53Seg~\cite{semanticKITTI} & \multicolumn{1}{l|}{\textbf{49.9}} & 86.4 & 24.5 & \textbf{32.7} & \textbf{25.5} & \textbf{22.6} & \textbf{36.2} & \textbf{33.6} &	4.7 & \textbf{91.8} & \textbf{64.8} & \textbf{74.6} &	\textbf{27.9} & \textbf{84.1} & \textbf{55.0} & 78.3 & \textbf{50.1} &	\textbf{64.0} &	\textbf{38.9} &	\textbf{52.2} \\
			
			\bottomrule
		
		\end{tabular}
	}
	\caption{Segmentation results of backbone networks and backbones combined with ASAP module on the Semantic KITTI dataset \cite{semanticKITTI}. Metric is mean IoU. PNv2 represents PointNet++ \cite{PointNet++} and SSv2 represents SqueezeSegV2 \cite{SqueezeSegV2}. SAP-x and ASAP-x are defined in Section~\ref{TE}. All methods were trained on sequences 00 to 10, except for sequence 08 which is used as validation set. We also list the results of other published works to give the readers an overview of the performances on the dataset.}
	\label{tab:results}
\end{table*}

\subsection{Model Analysis}


\subsubsection{Ablation on Spatio-temporal Correlation Strategy}
\label{updatingExperiment}
In this section, we do ablation analysis on the two spatio-temporal correlation strategies and input sequence length. Without loss of generality, we use PointNet++~\cite{PointNet++} with ASAP-1.

The results are shown in Table~\ref{tab:ablation}. As we can see, STC (i) is worse. We suppose it's because the focal regions (regions within a radius of center points) keep changing in different frames, making it hard to achieve temporal consistency. We can also see that the results get improved as the sequence length increases, which shows the effectiveness of our ASAP module to utilize temporal information.

\begin{table*}[h] 
	\scriptsize\centering
	\setlength{\tabcolsep}{0.45mm}{
		\begin{tabular}{@{}ccccccccccccccccc@{}}
			\toprule
			 & \multicolumn{1}{l|}{ param} & \multicolumn{1}{c|}{ frame} & & \multicolumn{1}{l|}{} & \multicolumn{12}{c}{IoU} \\
			Approach & \multicolumn{1}{c|}{ (M)} & \multicolumn{1}{c|}{num} & mIoU & \multicolumn{1}{l|}{mAcc} & 
			Bldg & Road & Sdwlk & Fence & Vegitn & Pole & Car & T.Sign & Pdstr & Bicyc & Lane & T.light \\ 
			\midrule
			PointNet++ \cite{PointNet++} & \multicolumn{1}{c|}{0.88} & \multicolumn{1}{c|}{ 1 } & 79.35 & \multicolumn{1}{l|}{85.43} & 96.88 & 97.72 & 86.20 & 92.75 & 97.12 & 97.09 & 90.85 & 66.87 & 78.64 & 0.00 & 72.93 & 75.17  \\
			PNv2 with STC (i) & \multicolumn{1}{c|}{1.84} & \multicolumn{1}{c|}{ 3} & 81.78 & \multicolumn{1}{l|}{86.47} & 97.18 & 97.73 & 89.91 & 94.46 & 96.61 & 97.01 & 94.25 & 78.28 & 83.74 & 0.00 & 74.69 & \textbf{77.69}  \\
			PNv2 with STC (ii) & \multicolumn{1}{c|}{1.84} & \multicolumn{1}{c|}{ 3 } & 82.73 & \multicolumn{1}{l|}{87.02} & 97.67 & 98.15 & 89.85 & 95.50 & 97.12 & 97.59 & 94.90 & 80.97 & 86.08 & 0.00 & 74.66 & 77.51  \\
			PNv2 with STC (ii) & \multicolumn{1}{c|}{1.84} & \multicolumn{1}{c|}{ 4 } & 82.82 & \multicolumn{1}{l|}{86.71} & \textbf{98.01} & 98.33 & 92.14 & \textbf{95.54} & 99.12 & \textbf{97.69} & 95.65 & 81.62 & 84.84 & 0.00 & 74.91 & 76.04  \\
			PNv2 with STC (ii) & \multicolumn{1}{c|}{1.84} & \multicolumn{1}{c|}{ 5 } & 82.90 & \multicolumn{1}{l|}{\textbf{87.14}} & 97.90 & \textbf{98.36} & 92.05 & 95.43 & \textbf{99.16} & 97.51 & 95.21 & \textbf{82.27} & 84.03 & 0.00 & 75.69 & 77.15  \\
			PNv2 with STC (ii) & \multicolumn{1}{c|}{1.84} & \multicolumn{1}{c|}{ 6 } & \textbf{83.00} & \multicolumn{1}{l|}{87.00} & 97.63 & 98.34 & \textbf{92.49} & 94.85 & 97.38 & 97.53 & \textbf{95.76} & 80.65 & \textbf{87.48} & 0.00 & \textbf{77.07} & 76.81  \\
			\bottomrule
		\end{tabular}
	}
	\caption{Segmentation results on the Synthia dataset \cite{synthia}. The evaluation metrics are mean IoU and mean accuracy ($\%$). PNv2 represents PointNet++~\cite{PointNet++}.}
	\label{tab:ablation}
\end{table*}

\subsubsection{Improvements on Different Backbones}
As we can see in Table~\ref{tab:results}, the improvement of PointNet++ \cite{PointNet++} is much larger than that of SqueezeSegV2 \cite{SqueezeSegV2}. We assume that's because our ASAP module architecture is similar with PointNet++ \cite{PointNet++} but totally different from SqueezeSegV2 \cite{SqueezeSegV2}. SqueezeSegV2 \cite{SqueezeSegV2} converts the point cloud to a 2D image. To combine it with our ASAP module, we have to first reconstruct point cloud from the image and then project the point cloud back to 2D. The process can cause information loss. For example, according to Section~\ref{implementation}, we use a max pooling layer to reconstruct the point cloud so the coordinates of output points can come from different points. In other words, those output points may not exist in the original point cloud at all, thus reducing the performance to some extent. 

\subsubsection{Failed Cases}
As we can see in Table~\ref{tab:results}, the IoU results of class truck and pole with backbone SqueezeSegV2 \cite{SqueezeSegV2} drop after incorporating ASAP-1. We hypothesize it's because there's not enough fine-scale information since the query radius of ASAP-1 is relatively large (1 meter). To verify that, we use multi-scale grouping proposed by \cite{PointNet++} in the LSA block to extract center point features. As shown in Table~\ref{tab:results}, results of both classes get improved.

\section{Conclusions}

We run a pilot study towards segmenting dynamic point cloud sequences by proposing a flexible module called ASAP. Based on properties of point cloud sequences, we introduce a novel attentive temporal embedding layer to fuse the relatively  informative local features across frames. We further present a spatio-temporal correlation to efficiently exploit more structure information for embedding. We show how we can easily insert the proposed ASAP module into former static point cloud pipelines and achieve great improvements on large-scale datasets. We expect our ASAP module can benefit research on point cloud sequence segmentation and related studies.

\bibliography{egbib}

\begin{thebibliography}{39}
\providecommand{\natexlab}[1]{#1}
\providecommand{\url}[1]{\texttt{#1}}
\expandafter\ifx\csname urlstyle\endcsname\relax
  \providecommand{\doi}[1]{doi: #1}\else
  \providecommand{\doi}{doi: \begingroup \urlstyle{rm}\Url}\fi

\bibitem[Azam et~al.(2018)Azam, Munir, Rafique, Ko, Sheri, and
  Jeon]{azam2018object}
Shoaib Azam, Farzeen Munir, Aasim Rafique, YeongMin Ko, Ahmad~Muqeem Sheri, and
  Moongu Jeon.
\newblock Object modeling from 3d point cloud data for self-driving vehicles.
\newblock In \emph{2018 IEEE Intelligent Vehicles Symposium (IV)}, pages
  409--414. IEEE, 2018.

\bibitem[Behl et~al.(2019)Behl, Paschalidou, Donne, and Geiger]{PointFlowNet}
Aseem Behl, Despoina Paschalidou, Simon Donne, and Andreas Geiger.
\newblock Pointflownet: Learning representations for rigid motion estimation
  from point clouds.
\newblock In \emph{The IEEE Conference on Computer Vision and Pattern
  Recognition (CVPR)}, June 2019.

\bibitem[Behley et~al.(2019)Behley, Garbade, Milioto, Quenzel, Behnke,
  Stachniss, and Gall]{semanticKITTI}
J.~Behley, M.~Garbade, A.~Milioto, J.~Quenzel, S.~Behnke, C.~Stachniss, and
  J.~Gall.
\newblock {SemanticKITTI: A Dataset for Semantic Scene Understanding of LiDAR
  Sequences}.
\newblock In \emph{Proc. of the IEEE/CVF International Conf.~on Computer Vision
  (ICCV)}, 2019.

\bibitem[Biasutti et~al.(2019)Biasutti, Bugeau, Aujol, and Br{\'e}dif]{RIUNet}
Pierre Biasutti, Aur{\'e}lie Bugeau, Jean-Fran{\c{c}}ois Aujol, and Mathieu
  Br{\'e}dif.
\newblock Riu-net: Embarrassingly simple semantic segmentation of 3d lidar
  point cloud.
\newblock \emph{arXiv:1905.08748}, 2019.

\bibitem[Choy et~al.(2019)Choy, Gwak, and Savarese]{4DConv}
Christopher Choy, JunYoung Gwak, and Silvio Savarese.
\newblock 4d spatio-temporal convnets: Minkowski convolutional neural networks.
\newblock In \emph{Proceedings of the IEEE Conference on Computer Vision and
  Pattern Recognition}, pages 3075--3084, 2019.

\bibitem[Dewan et~al.(2016)Dewan, Caselitz, Tipaldi, and
  Burgard]{RigidSceneFlow}
Ayush Dewan, Tim Caselitz, Gian~Diego Tipaldi, and Wolfram Burgard.
\newblock Rigid scene flow for 3d lidar scans.
\newblock In \emph{2016 IEEE/RSJ International Conference on Intelligent Robots
  and Systems (IROS)}, pages 1765--1770. IEEE, 2016.

\bibitem[Ding et~al.(2019)Ding, Terwilliger, Sherony, Reimer, and
  Fridman]{ValueOfTem}
Li~Ding, Jack Terwilliger, Rini Sherony, Bryan Reimer, and Lex Fridman.
\newblock Value of temporal dynamics information in driving scene segmentation.
\newblock \emph{arXiv:1904.00758}, 2019.

\bibitem[Engelmann et~al.(2018)Engelmann, Kontogianni, Schult, and
  Leibe]{KnowNeighbors}
Francis Engelmann, Theodora Kontogianni, Jonas Schult, and Bastian Leibe.
\newblock Know what your neighbors do: 3d semantic segmentation of point
  clouds.
\newblock In \emph{The European Conference on Computer Vision (ECCV)
  Workshops}, September 2018.

\bibitem[Graham et~al.(2018)Graham, Engelcke, and van~der Maaten]{SSCNN}
Benjamin Graham, Martin Engelcke, and Laurens van~der Maaten.
\newblock 3d semantic segmentation with submanifold sparse convolutional
  networks.
\newblock In \emph{The IEEE Conference on Computer Vision and Pattern
  Recognition (CVPR)}, June 2018.

\bibitem[Groh et~al.(2018)Groh, Wieschollek, and Lensch]{FlexConvolution}
Fabian Groh, Patrick Wieschollek, and Hendrik Lensch.
\newblock Flex-convolution (million-scale point-cloud learning beyond
  grid-worlds).
\newblock \emph{arXiv:1803.07289}, 2018.

\bibitem[Hua et~al.(2018)Hua, Tran, and Yeung]{PointwiseConv}
Binh-Son Hua, Minh-Khoi Tran, and Sai-Kit Yeung.
\newblock Pointwise convolutional neural networks.
\newblock In \emph{The IEEE Conference on Computer Vision and Pattern
  Recognition (CVPR)}, June 2018.

\bibitem[Huang and You(2016)]{PointLabeling}
Jing Huang and Suya You.
\newblock Point cloud labeling using 3d convolutional neural network.
\newblock In \emph{2016 23rd International Conference on Pattern Recognition
  (ICPR)}, pages 2670--2675. IEEE, 2016.

\bibitem[Iandola et~al.(2016)Iandola, Han, Moskewicz, Ashraf, Dally, and
  Keutzer]{SqueezeNet}
Forrest~N Iandola, Song Han, Matthew~W Moskewicz, Khalid Ashraf, William~J
  Dally, and Kurt Keutzer.
\newblock Squeezenet: Alexnet-level accuracy with 50x fewer parameters and< 0.5
  mb model size.
\newblock \emph{arXiv:1602.07360}, 2016.

\bibitem[Jiang et~al.(2018)Jiang, Wu, Zhao, Zhao, and Lu]{PointSIFT}
Mingyang Jiang, Yiran Wu, Tianqi Zhao, Zelin Zhao, and Cewu Lu.
\newblock Pointsift: A sift-like network module for 3d point cloud semantic
  segmentation.
\newblock \emph{arXiv:1807.00652}, 2018.

\bibitem[Klokov and Lempitsky(2017)]{EscapeFromCells}
Roman Klokov and Victor Lempitsky.
\newblock Escape from cells: Deep kd-networks for the recognition of 3d point
  cloud models.
\newblock In \emph{The IEEE International Conference on Computer Vision
  (ICCV)}, Oct 2017.

\bibitem[Kung et~al.(2016)Kung, Huang, and Chien]{kung2016efficient}
Yen-Cheng Kung, Yung-Lin Huang, and Shao-Yi Chien.
\newblock Efficient surface detection for augmented reality on 3d point clouds.
\newblock In \emph{Proceedings of the 33rd Computer Graphics International},
  pages 89--92. 2016.

\bibitem[Landrieu and Simonovsky(2018)]{SPGraph}
Loic Landrieu and Martin Simonovsky.
\newblock Large-scale point cloud semantic segmentation with superpoint graphs.
\newblock In \emph{The IEEE Conference on Computer Vision and Pattern
  Recognition (CVPR)}, June 2018.

\bibitem[Liu et~al.(2019{\natexlab{a}})Liu, Qi, and Guibas]{FlowNet3D}
Xingyu Liu, Charles~R. Qi, and Leonidas~J. Guibas.
\newblock Flownet3d: Learning scene flow in 3d point clouds.
\newblock In \emph{The IEEE Conference on Computer Vision and Pattern
  Recognition (CVPR)}, June 2019{\natexlab{a}}.

\bibitem[Liu et~al.(2019{\natexlab{b}})Liu, Yan, and Bohg]{meteornet}
Xingyu Liu, Mengyuan Yan, and Jeannette Bohg.
\newblock Meteornet: Deep learning on dynamic 3d point cloud sequences.
\newblock In \emph{Proceedings of the IEEE International Conference on Computer
  Vision}, pages 9246--9255, 2019{\natexlab{b}}.

\bibitem[Pan and Bingham(2013)]{lowvision}
Jing~Samantha Pan and Geoffrey~P Bingham.
\newblock With an eye to low vision: Optic flow enables perception despite
  image blur.
\newblock \emph{Optometry and Vision Science}, 90\penalty0 (10):\penalty0
  1119--1127, 2013.

\bibitem[Pang et~al.(2019)Pang, Zha, Cao, Shi, and Lu]{deepRNN}
Bo~Pang, Kaiwen Zha, Hanwen Cao, Chen Shi, and Cewu Lu.
\newblock Deep rnn framework for visual sequential applications.
\newblock In \emph{Proceedings of the IEEE conference on computer vision and
  pattern recognition}, pages 423--432, 2019.

\bibitem[Pang et~al.(2020)Pang, Zha, Cao, Tang, Yu, and Lu]{nature}
Bo~Pang, Kaiwen Zha, Hanwen Cao, Jiajun Tang, Minghui Yu, and Cewu Lu.
\newblock Complex sequential understanding through the awareness of spatial and
  temporal concepts.
\newblock \emph{Nature Machine Intelligence}, 2\penalty0 (5):\penalty0
  245--253, 2020.

\bibitem[Qi et~al.(2017{\natexlab{a}})Qi, Su, Mo, and Guibas]{PointNet}
Charles~R. Qi, Hao Su, Kaichun Mo, and Leonidas~J. Guibas.
\newblock Pointnet: Deep learning on point sets for 3d classification and
  segmentation.
\newblock In \emph{The IEEE Conference on Computer Vision and Pattern
  Recognition (CVPR)}, July 2017{\natexlab{a}}.

\bibitem[Qi et~al.(2017{\natexlab{b}})Qi, Yi, Su, and Guibas]{PointNet++}
Charles~Ruizhongtai Qi, Li~Yi, Hao Su, and Leonidas~J Guibas.
\newblock Pointnet++: Deep hierarchical feature learning on point sets in a
  metric space.
\newblock In \emph{Advances in neural information processing systems}, pages
  5099--5108, 2017{\natexlab{b}}.

\bibitem[Riegler et~al.(2017)Riegler, Osman~Ulusoy, and Geiger]{OctNet}
Gernot Riegler, Ali Osman~Ulusoy, and Andreas Geiger.
\newblock Octnet: Learning deep 3d representations at high resolutions.
\newblock In \emph{The IEEE Conference on Computer Vision and Pattern
  Recognition (CVPR)}, July 2017.

\bibitem[Ronneberger et~al.(2015)Ronneberger, Fischer, and Brox]{UNet}
Olaf Ronneberger, Philipp Fischer, and Thomas Brox.
\newblock U-net: Convolutional networks for biomedical image segmentation.
\newblock In \emph{International Conference on Medical image computing and
  computer-assisted intervention}, pages 234--241. Springer, 2015.

\bibitem[Ros et~al.(2016)Ros, Sellart, Materzynska, Vazquez, and
  Lopez]{synthia}
German Ros, Laura Sellart, Joanna Materzynska, David Vazquez, and Antonio~M.
  Lopez.
\newblock The synthia dataset: A large collection of synthetic images for
  semantic segmentation of urban scenes.
\newblock In \emph{The IEEE Conference on Computer Vision and Pattern
  Recognition (CVPR)}, June 2016.

\bibitem[Salah et~al.(2017)Salah, Kramm, Demonceaux, and
  Vasseur]{salah2017summarizing}
Imeen~Ben Salah, Sebastien Kramm, C{\'e}dric Demonceaux, and Pascal Vasseur.
\newblock Summarizing large scale 3d point cloud for navigation tasks.
\newblock In \emph{2017 IEEE 20th International Conference on Intelligent
  Transportation Systems (ITSC)}, pages 1--8. IEEE, 2017.

\bibitem[Su et~al.(2018)Su, Jampani, Sun, Maji, Kalogerakis, Yang, and
  Kautz]{SPLATNet}
Hang Su, Varun Jampani, Deqing Sun, Subhransu Maji, Evangelos Kalogerakis,
  Ming-Hsuan Yang, and Jan Kautz.
\newblock Splatnet: Sparse lattice networks for point cloud processing.
\newblock In \emph{The IEEE Conference on Computer Vision and Pattern
  Recognition (CVPR)}, June 2018.

\bibitem[Tatarchenko et~al.(2018)Tatarchenko, Park, Koltun, and
  Zhou]{TangentConv}
Maxim Tatarchenko, Jaesik Park, Vladlen Koltun, and Qian-Yi Zhou.
\newblock Tangent convolutions for dense prediction in 3d.
\newblock In \emph{The IEEE Conference on Computer Vision and Pattern
  Recognition (CVPR)}, June 2018.

\bibitem[Tchapmi et~al.(2017)Tchapmi, Choy, Armeni, Gwak, and
  Savarese]{SEGCloud}
Lyne Tchapmi, Christopher Choy, Iro Armeni, JunYoung Gwak, and Silvio Savarese.
\newblock Segcloud: Semantic segmentation of 3d point clouds.
\newblock In \emph{2017 international conference on 3D vision (3DV)}, pages
  537--547. IEEE, 2017.

\bibitem[Te et~al.(2018)Te, Hu, Zheng, and Guo]{RGCNN}
Gusi Te, Wei Hu, Amin Zheng, and Zongming Guo.
\newblock Rgcnn: Regularized graph cnn for point cloud segmentation.
\newblock In \emph{Proceedings of the 26th ACM international conference on
  Multimedia}, pages 746--754, 2018.

\bibitem[Tombari et~al.(2010)Tombari, Salti, and Di~Stefano]{SHOT}
Federico Tombari, Samuele Salti, and Luigi Di~Stefano.
\newblock Unique signatures of histograms for local surface description.
\newblock In \emph{European conference on computer vision}, pages 356--369.
  Springer, 2010.

\bibitem[Ushani et~al.(2017)Ushani, Wolcott, Walls, and
  Eustice]{LearnigApproachSceneFlow}
Arash~K Ushani, Ryan~W Wolcott, Jeffrey~M Walls, and Ryan~M Eustice.
\newblock A learning approach for real-time temporal scene flow estimation from
  lidar data.
\newblock In \emph{2017 IEEE International Conference on Robotics and
  Automation (ICRA)}, pages 5666--5673. IEEE, 2017.

\bibitem[Wang et~al.(2019)Wang, Sun, Liu, Sarma, Bronstein, and
  Solomon]{DynamicGraph}
Yue Wang, Yongbin Sun, Ziwei Liu, Sanjay~E Sarma, Michael~M Bronstein, and
  Justin~M Solomon.
\newblock Dynamic graph cnn for learning on point clouds.
\newblock \emph{ACM Transactions on Graphics (TOG)}, 38\penalty0 (5):\penalty0
  1--12, 2019.

\bibitem[Wang and Lu(2019)]{VoxSegNet}
Zongji Wang and Feng Lu.
\newblock Voxsegnet: Volumetric cnns for semantic part segmentation of 3d
  shapes.
\newblock \emph{IEEE transactions on visualization and computer graphics},
  2019.

\bibitem[Wu et~al.(2018)Wu, Wan, Yue, and Keutzer]{SqueezeSeg}
Bichen Wu, Alvin Wan, Xiangyu Yue, and Kurt Keutzer.
\newblock Squeezeseg: Convolutional neural nets with recurrent crf for
  real-time road-object segmentation from 3d lidar point cloud.
\newblock In \emph{2018 IEEE International Conference on Robotics and
  Automation (ICRA)}, pages 1887--1893. IEEE, 2018.

\bibitem[Wu et~al.(2019)Wu, Zhou, Zhao, Yue, and Keutzer]{SqueezeSegV2}
Bichen Wu, Xuanyu Zhou, Sicheng Zhao, Xiangyu Yue, and Kurt Keutzer.
\newblock Squeezesegv2: Improved model structure and unsupervised domain
  adaptation for road-object segmentation from a lidar point cloud.
\newblock In \emph{2019 International Conference on Robotics and Automation
  (ICRA)}, pages 4376--4382. IEEE, 2019.

\bibitem[Zeng and Gevers(2018)]{KdTreeGuided}
Wei Zeng and Theo Gevers.
\newblock 3dcontextnet: K-d tree guided hierarchical learning of point clouds
  using local and global contextual cues.
\newblock In \emph{The European Conference on Computer Vision (ECCV)
  Workshops}, September 2018.

\end{thebibliography}
\end{document}